\documentclass[letterpaper, 10 pt, conference]{ieeeconf}  
\IEEEoverridecommandlockouts                              
\usepackage{amsmath}
\usepackage{amssymb}
\usepackage{gensymb}
\usepackage{algorithm} 
\usepackage{algpseudocode}
\usepackage{url}
\usepackage{graphicx}
\usepackage{float}
\usepackage{setspace}
\usepackage{subfigure}
\usepackage{xcolor}
\usepackage{wrapfig}
\usepackage{makecell}

\renewcommand{\baselinestretch}{0.965}

\newtheorem{remark}{Remark}

\title{\LARGE \bf
Generating Out-Of-Distribution Scenarios Using Language Models}

\author{Erfan Aasi$^{1}$, Phat Nguyen$^{2}$, Shiva Sreeram$^{1}$, Guy Rosman$^{3}$, Sertac Karaman$^{4}$, and Daniela Rus$^{1}$
\thanks{*This work is supported by Toyota Research Institute (TRI). It, however, reflects solely the opinions and conclusions of its authors, and not TRI or any other Toyota entity.}
\thanks{$^{1}$MIT CSAIL, $^{2}$UMass Amherst, $^{3}$TRI, $^{4}$MIT LIDS}
}

\begin{document}

\maketitle
\thispagestyle{empty}
\pagestyle{empty}

%%%%%%%%%%%%%%%%%%%%%%%%%%%%%%%%%%%%%%%%%%%%%%%%%%%%%%%%%%%%%%%
%%%%%%%%%%%%%%%%%%%%%%%%%%% ABSTRACT %%%%%%%%%%%%%%%%%%%%%%%%%%
\begin{abstract}
The deployment of autonomous vehicles controlled by machine learning techniques requires extensive testing in diverse real-world environments, robust handling of edge cases and out-of-distribution scenarios, and comprehensive safety validation to ensure that these systems can navigate safely and effectively under unpredictable conditions. Addressing Out-Of-Distribution (OOD) driving scenarios is essential for enhancing safety, as OOD scenarios help validate the reliability of the models within the vehicle’s autonomy stack. However, generating OOD scenarios is challenging due to their long-tailed distribution and rarity in urban driving datasets. Recently, Large Language Models (LLMs) have shown promise in autonomous driving, particularly for their zero-shot generalization and common-sense reasoning capabilities. In this paper, we leverage these LLM strengths to introduce a framework for generating diverse OOD driving scenarios. Our approach uses LLMs to construct a branching tree, where each branch represents a unique OOD scenario. These scenarios are then simulated in the CARLA simulator using an automated framework that aligns scene augmentation with the corresponding textual descriptions.
We evaluate our framework through extensive simulations, and assess its performance via a diversity metric that measures the richness of the scenarios. Additionally, we introduce a new "OOD-ness" metric, which quantifies how much the generated scenarios deviate from typical urban driving conditions. Furthermore, we explore the capacity of modern Vision-Language Models (VLMs) to interpret and safely navigate through the simulated OOD scenarios. Our findings offer valuable insights into the reliability of language models in addressing OOD scenarios within the context of urban driving.
\end{abstract}

%%%%%%%%%%%%%%%%%%%%%%%%%%%%%%%%%%%%%%%%%%%%%%%%%%%%%%%%%%%%%%%
%%%%%%%%%%%%%%%%%%%%%%%%%%% INTRODUCTION %%%%%%%%%%%%%%%%%%%%%%
\section{INTRODUCTION} \label{sec:introduction}
% First Paragraph: AVs are getting better but still struggling with OODs (short definition of OOD), especially as the modern ones are data-driven 

\begin{figure} [htb]
    \centering
    \subfigure[]
    {\includegraphics[width=0.49\columnwidth]{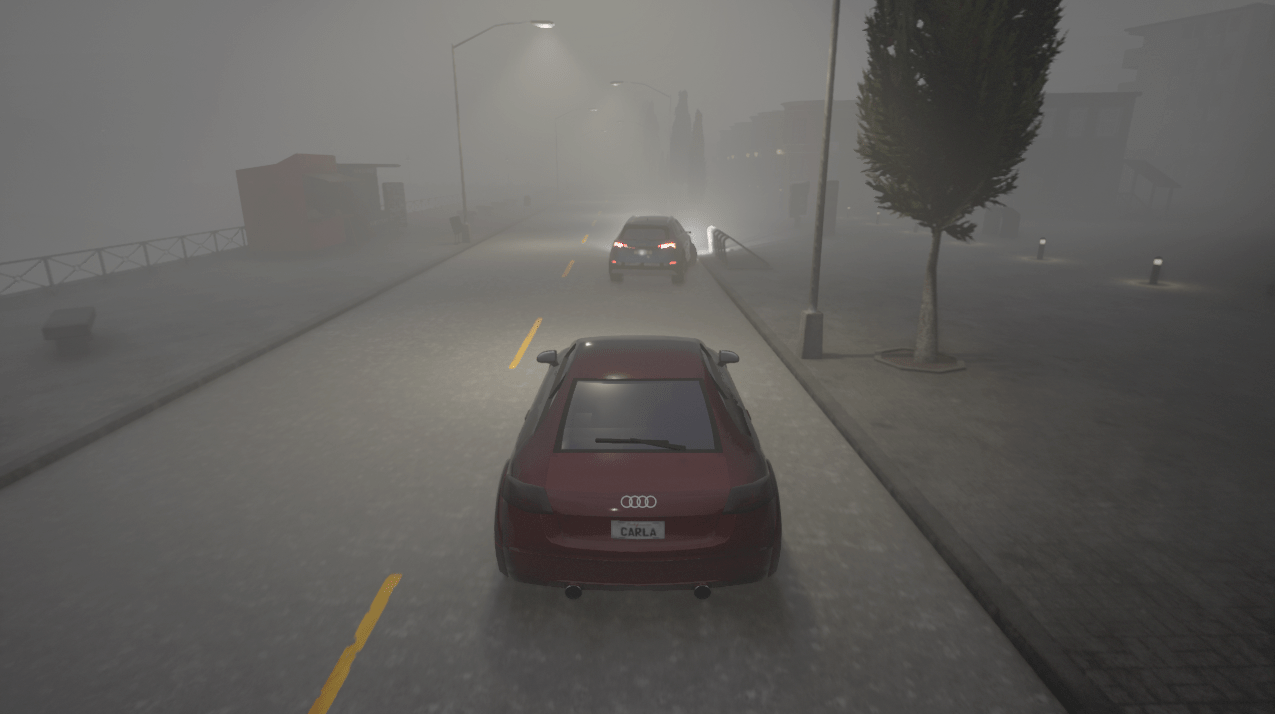}\label{fig:dense_fog}}
    \subfigure[]
    {\includegraphics[width=0.49\columnwidth]{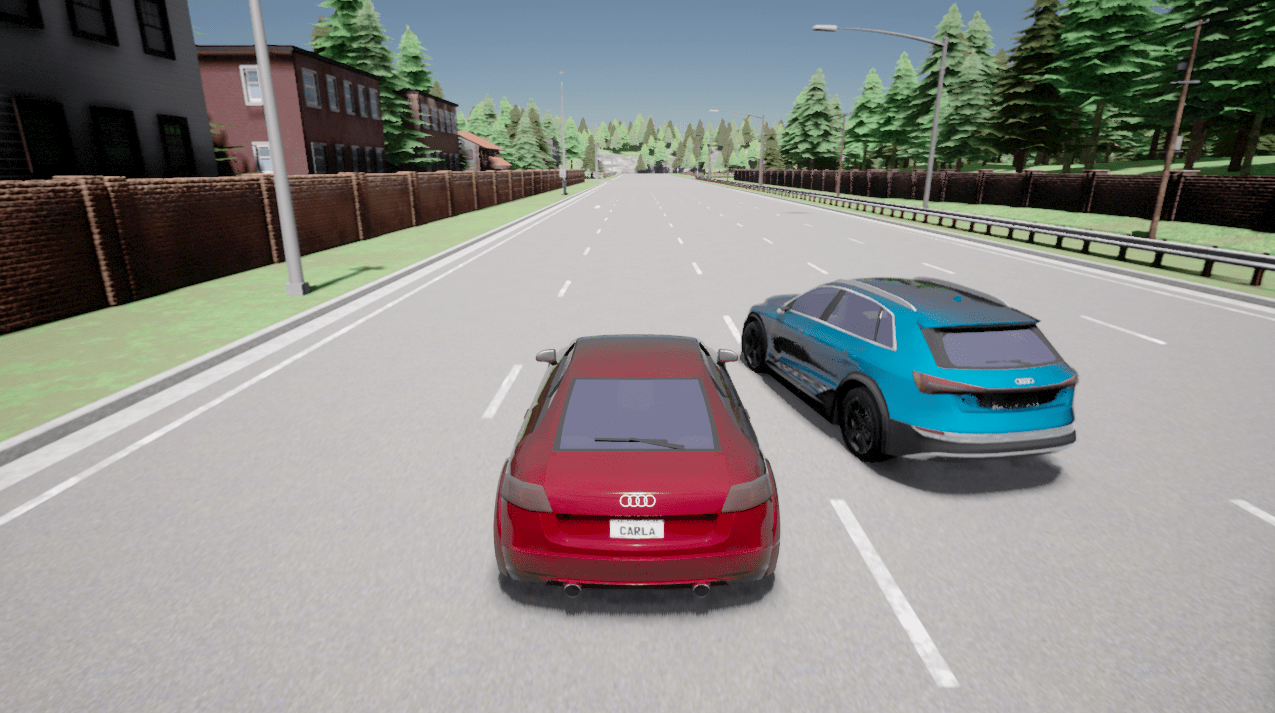}\label{fig:lane_change}}
    \caption{Snapshots of OOD scenarios implemented in the CARLA simulator \cite{dosovitskiy2017carla}. In both instances, the ego vehicle, represented by the red Audi, encounters challenging situations: (a) navigating through dense fog with restricted visibility, and (b) driving on a highway when a nearby vehicle abruptly switches into the ego's lane.}
\label{fig:ood_examples}
\vspace{-4mm}
\end{figure}

Autonomous driving has seen significant progress in recent years, primarily through AI-based end-to-end machine learning approaches \cite{shao2024lmdrive, chen2024driving, wang2024drive, amini2018variational}. These data-driven methods aim to capture and model the underlying distributions within the collected data. However, given the complexity of autonomous driving systems and the broad spectrum of real-world scenarios, these models often struggle to fully account for the vast variability of the real world, leaving them potentially vulnerable to Out-Of-Distribution (OOD) data—data that significantly differs from what the model was trained on (see Fig.~\ref{fig:ood_examples})\cite{salehi2021unified, bogdoll2021description, zhou2023corner, jain2021autonomy}. While self-driving cars are demonstrating promising performance in closed environments with in-distribution and well-behaved scenarios, achieving practical autonomous driving and ensuring safe navigation in real-world environments requires autonomous vehicles to handle a wide range of OOD scenarios.

% Second Paragraph: Automatically generating OODs is crucial + some literature works are aiming on it, but they lack diversity or automation

The infrequent occurrence of OOD scenarios in the real world presents significant challenges for the safe and effective deployment of an autonomous vehicle, often referred to as the \textit{ego} vehicle, in urban settings.  The vast number of potential scenario variations, driven by the combinatorial possibilities of different interacting elements, makes it computationally infeasible to thoroughly explore all possible scenarios to identify OOD cases \cite{wang2021advsim}. As a result, the automatic generation of diverse OOD scenarios is essential for tackling the long-tail issue in autonomous driving \cite{liu2024safety}.

% Third Paragraph: LLMs are great in autonomous driving + intro to our method

In this paper we present a framework for generating diverse OOD driving scenarios, using language models. Foundation models, with their contextual reasoning and zero-shot generalization abilities, are a promising candidate for identifying OOD scenarios in autonomous systems \cite{sinha2024real, li2024automated}. Our approach combines the GPT-4o model \cite{gpt4o} with few-shot Chain-of-Thought (CoT) prompting \cite{brown2020language,wei2022chain} to generate a tree, where each branch represents a distinct OOD scenario. This initial tree is subsequently refined through language model-based red teaming, utilizing a red LLM \cite{perez2022red,hong2024curiosity}. This refinement enhances both  the diversity and quality of the scenarios, resulting in what we referred to as the \textit{diverse tree}. The diverse tree is used to generate textual descriptions of various OOD scenarios, providing a broad range of situations for the autonomous vehicles to explore.

% Fourth Paragraph: Our method

Our method simulates the OOD scenarios using the CARLA simulator \cite{dosovitskiy2017carla}. Based on the available assets in CARLA, the diverse tree is pruned to form a \textit{simulatable tree}, which is used to generate textual descriptions of OOD scenarios that can be simulated in CARLA. To automate the simulation process, we employ another LLM to interpret the textual OOD descriptions and suggest appropriate states and behaviors for the objects involved in each scenario. To evaluate the performance of our framework, specifically the quality of the generated scenarios, we define two key metrics: OOD-ness and diversity. OOD-ness quantifies how far the generated scenarios deviate from typical urban driving conditions, while diversity measures the variability of the scenarios and their ability to capture different situations. Comparisons with baseline datasets, e.g., nuScenes \cite{caesar2020nuscenes}, highlights the effectiveness of our approach in generating diverse OOD scenarios. Finally, we assess the performance of modern Vision Language Models (VLMs) on our generated OOD scenarios to evaluate their reliability in identifying these scenarios as OODs and selecting the appropriate safe actions. This provides a valuable benchmark for determining the trustworthiness of VLMs within autonomy stacks.

% Fifth Paragraph: Contributions

We summarize our major contributions as follows:
\begin{itemize}
    \item We provide a language-model based framework for generating textual descriptions of diverse OOD driving scenarios. A tree structure is developed to produce a variety of OOD descriptions, with each path along the tree representing a unique OOD scenario. We introduce the OOD-ness and diversity metrics for assessing the quality of the generated scenarios. 

    \item We create an automated pipeline for simulating OOD scenarios, by seamlessly translating the textual descriptions of these scenarios into their corresponding simulations in CARLA.

    \item We assess the zero-shot capabilities of state-of-the-art VLMs on the simulated OOD scenarios, focusing on their ability to recognize the OOD nature of these scenarios and determine the safest control actions to navigate through them.
\end{itemize}

%%%%%%%%%%%%%%%%%%%%%%%%%%%%%%%%%%%%%%%%%%%%%%%%%%%%%%%%%%%%%%%
%%%%%%%%%%%%%%%%%%%%%%%%%%% RELATED WORKS %%%%%%%%%%%%%%%%%%%%%
\section{RELATED WORKS} \label{sec:related_works}

\subsection{OOD Scenarios in Autonomous Driving} \label{subsec:oods_in_ad}
The study of OOD cases has attracted significant attention in recent years within the fields of machine learning and robotic, particularly in the context of autonomous driving systems \cite{liu2019large,makansi2021exposing,bogdoll2021description,zhou2023corner}. Approaches for addressing OOD scenarios can be classified into three distinct categories:

\subsubsection{Robustness vs OODs} these methods focus on strengthening a model's performance when confronted with OOD scenarios. For instance, in \cite{stoler2024safeshift}, they propose a trajectory prediction approach under a safety-informed distribution shift setting. Similarly, in \cite{li2024adaptive} they introduced a framework for trajectory prediction by integrating deep learning-based and rule-based prediction models, aimed to improve the generalization capability of trajectory prediction models to OOD scenario. In \cite{arasteh2024learning}, an imitation learning-based framework is presented to mitigate OOD cases in the planning level.

\subsubsection{OOD detection} these methods aim to identify when the data provided to an autonomous system is different from the training data, primarily in the context of perception in autonomous driving. For example, in \cite{bolte2019towards}, they propose a framework to detect such scenarios in both offline and online settings, using input video frames. An end-to-end approach for online OOD detection in the perception level is introduced in \cite{kaljavesi2024integrating}, while \cite{bogdoll2024hybrid} presents a hybrid video anomaly detection framework. Additionally, \cite{nitsch2021out} develops a generative adversarial network-based framework for OOD detection, without requiring OOD data during training.

\subsubsection{OOD generation} these studies focus on generating OOD scenarios, which provide valuable datasets for assessing the reliability of self-driving cars' autonomy stacks. This category is the primary focus of our paper. In \cite{wang2021advsim}, an adversarial framework is proposed to generate safety-critical scenarios by perturbing the maneuvers of interactive actors through adversarial behaviors. The CODA dataset, introduced in \cite{li2022coda}, serves as a corner case dataset for autonomous driving, based on real-world road data from KITTI \cite{geiger2012we}, nuScenes \cite{caesar2020nuscenes}, and ONCE \cite{mao2021one}. A deep reinforcement learning-based method for generating safety-critical scenarios is presented in \cite{liu2024safety}, which involves sequential editing actions such as adding new agents or altering the trajectories of existing ones. Finally, \cite{hao2023adversarial} introduces a safety-critical scenario generation framework that uses generative adversarial imitation learning to develop a human driving prior model, subsequently employed as a reward function in a reinforcement learning-based framework to generate adversarial scenarios. Although these methods show promising performance in generating safety-critical driving scenarios, they primarily focus on vehicle-to-vehicle interactions and do not cover the full spectrum of OOD scenarios.

\subsection{Language Models in Autonomous Driving} \label{subsec:lms_in_ad}

Recently, LLMs have been incorporated into driving systems to harness their contextual understanding and common-sense reasoning capabilities, particularly for addressing long-tail scenarios. Knowledge-driven approaches such as Dilu \cite{wen2023dilu} and Drive Like a Human \cite{fu2024drive} are porposed as decision makers for autonomous driving. LLM-based driving frameworks such as DriveLM \cite{sima2023drivelm}, DriveGPT4 \cite{xu2024drivegpt4}, Text-to-Drive \cite{nguyen2024text}, LLM-Driver \cite{chen2024driving}, and DriveMLM \cite{wang2023drivemlm} utilize multimodal inputs for driving. End-to-end language model-based frameworks such as LMDrive \cite{shao2024lmdrive} and Drive Anywhere \cite{wang2024drive} have also been proposed for autonomous driving. The encouraging results achieved by LLMs in autonomous driving inspire us to integrate them into our framework in this paper.

\subsection{Language Models to Address OODs} \label{subsec:lms_for_oods}

The promising performance of LLMs has inspired several recent studies to tackle OOD scenarios. In \cite{sinha2024real}, a real-time LLM-based framework is proposed to detect anomalous behaviors in autonomous systems and determine appropriate safety-preserving actions. In \cite{elhafsi2023semantic}, they provide a monitoring framework for the semantic detection of OOD scenarios in vision-based policies for autonomous systems. In \cite{li2024automated}, an approach for the automatic evaluation of VLMs on OOD scenarios is proposed, using the CODA dataset \cite{li2022coda}. The closest work to our paper is \cite{lu2024multimodal}, which focuses on using LLMs to generate corner cases by designing road structures and vehicle-to-vehicle interactions. However, this approach does not address the broad spectrum of OOD scenarios in autonomous driving.

%%%%%%%%%%%%%%%%%%%%%%%%%%%%%%%%%%%%%%%%%%%%%%%%%%%%%%%%%%%%%%%%%%%%%%%%%
%%%%%%%%%%%%%%%%%%%%% METHODOLOGY %%%%%%%%%%%%%%%%%%%%%%%%%%%%%%%%%%%%%%%
\section{METHODOLOGY} \label{section:methodology}

\subsection{Diverse OOD Scenario Generation} \label{section:diverse_scenario_generation}
Although there is no clear definition of OOD driving scenarios \cite{salehi2021unified,bogdoll2021description}, in this paper we consider them as rare cases that are not usually encountered in urban driving environments and therefore may not be used in training the autonomous driving algorithms. To develop a comprehensive framework that captures a diverse set of OOD driving scenarios, and by drawing inspiration from accident report data \cite{national2008national}, we classify these scenarios into two groups:

\subsubsection{Environmental OOD Scenarios}
It involves situations where an environmental element significantly deviates from its usual state and affects the performance of ego. These scenarios do not typically involve dynamic interactions with other road users but arise from anomalies in the surrounding environment that ego must detect and respond to appropriately. Examples include extreme weather conditions such as heavy snowfall or dense fog, or unexpected road obstructions caused by static objects such as a fallen tree or large debris. 

\subsubsection{Interactional OOD Scenarios}
It involves situations where ego encounters an unexpected or unusual interaction with another road user or dynamic actor. These scenarios require ego to actively engage in real-time decision making and interaction, often involving sudden or unpredictable behavior from other dynamic objects that falls outside the pre-defined interaction patterns.  Examples include a pedestrian suddenly crossing the road in an unexpected location or an animal, like a deer or dog, suddenly appearing on the road.

\begin{figure}[htb]
\centerline{\includegraphics[width=0.8\columnwidth]{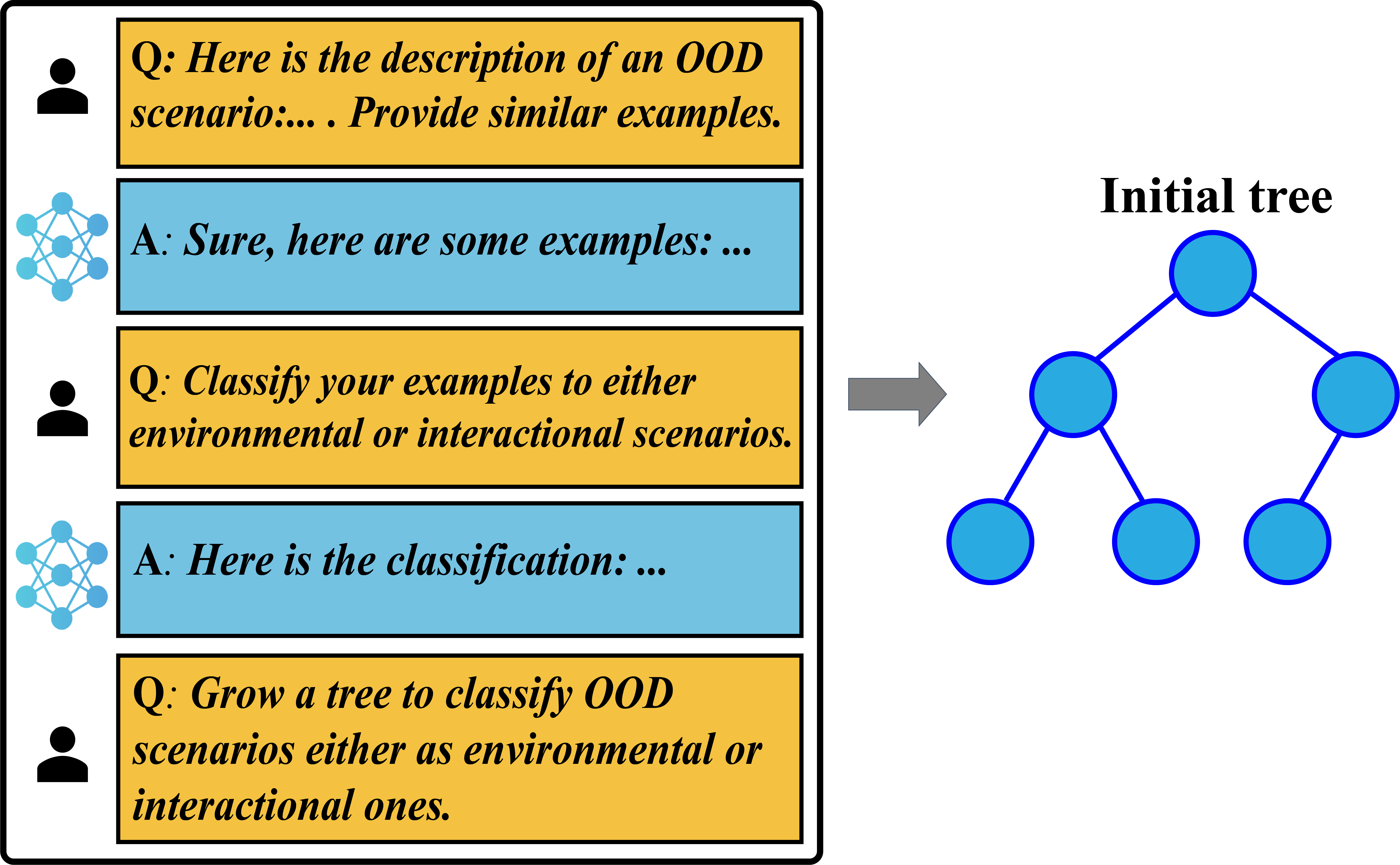}}
\caption{Illustration of employing an LLM in a few-shot CoT approach to construct an initial tree for generating OOD scenarios.}
\label{fig:initial_tree}
\end{figure}

To create a diverse set of OOD scenarios, we utilize LLMs through a few-shot CoT prompting approach (see Fig.~\ref{fig:initial_tree}). We begin by introducing the concept of an OOD driving scenario to the LLM and requesting it to generate examples of similar scenarios. Next, we present the classification pattern we use for OOD scenarios and prompt the LLM to categorize its generated examples into one of the classes. Finally, inspired by the idea of \cite{yao2024tree}, we instruct the LLM to generate a tree structure that can be used to produce OOD scenarios, where each path in the tree corresponds to a distinct OOD description. The LLM's output provides an \textit{initial tree} for defining OOD scenarios. We refer to this LLM as \textit{tree-LLM}.

\begin{figure}[htb]
\centerline{\includegraphics[width=1.0\columnwidth]{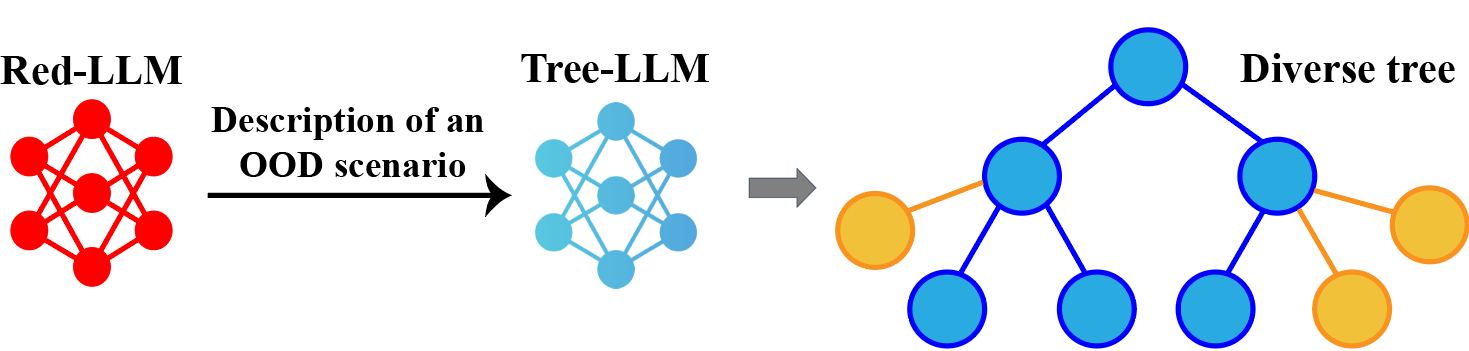}}
\caption{Employing the red-LLM to enrich the diversity of the initial tree, which leads to the diverse tree.}
\label{fig:diverse_tree}
\end{figure}

Later to enrich the diversity of initial tree, we employ an idea inspired by LLM-based red teaming technique (see Fig.~\ref{fig:diverse_tree}). We prompt an LLM, referred to as \textit{red-LLM}, using a few-shot approach to generate textual descriptions of potential OOD scenarios that could occur in urban driving environments. The textual output of the red-LLM is subsequently fed into the tree-LLM, which is instructed either to identify an existing branch in the initial tree that represents the given OOD scenario or to modify the tree structure to accommodate it. This refinement is achieved by either adjusting existing nodes to cover a broader range of elements or by adding new nodes to the tree. This open-loop process can be repeated multiple times, resulting in a \textit{diverse tree} that encompasses a wider range of OOD scenarios.

\subsection{Automated Simulation of OOD Scenarios} \label{section:automated_simulation}
\begin{figure}[htb]
\centerline{\includegraphics[width=0.7\columnwidth]{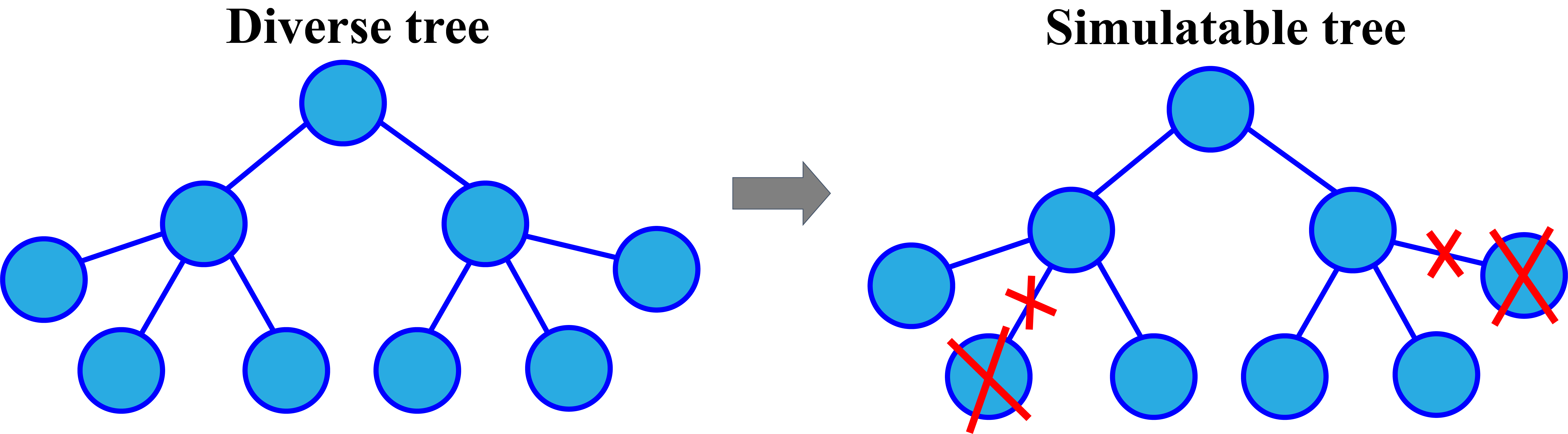}}
\caption{Pruning the diverse tree to a simulatable tree, based on the available assets in CARLA.}
\label{fig:simulatable_tree}
\end{figure}

To fully leverage the benefits of generating OOD scenarios, it is essential to implement them in a simulator and evaluate the autonomy stacks of self-driving cars in a simulated environment before deploying them in real-world vehicles. Note that the OOD scenarios generated by the tree-LLM are synthetic and may involve vast number of elements or extremely complicated interactions. However, every simulator has its own limitations and resources, and not all the OOD scenarios generated by the diverse tree are necessarily simulatable. To address this concern, we first prune the diverse tree based on the available assets in CARLA, which leads to a diverse tree that is also simulatable, referred to as \textit{simulatable tree} (see Fig.~\ref{fig:simulatable_tree}).

We then prompt the tree-LLM to generate textual descriptions of OOD scenarios, based on the structure of the simulatable tree. To automate the process of connecting these textual descriptions to actual simulation traces, we use another LLM, referred to as \textit{Augmenter-LLM}. This LLM takes the textual description of the OOD scenario, along with the available assets in CARLA, and is tasked with providing the augmentation details of the scene to align with the scenario's description. This process involves selecting an appropriate map in the simulator, adjusting the weather conditions, determining the spawning position of objects (relatively with respect to ego's position), and deciding the motion patterns of dynamic objects.

\begin{figure}[htb]
\centerline{\includegraphics[width=1.0\columnwidth]{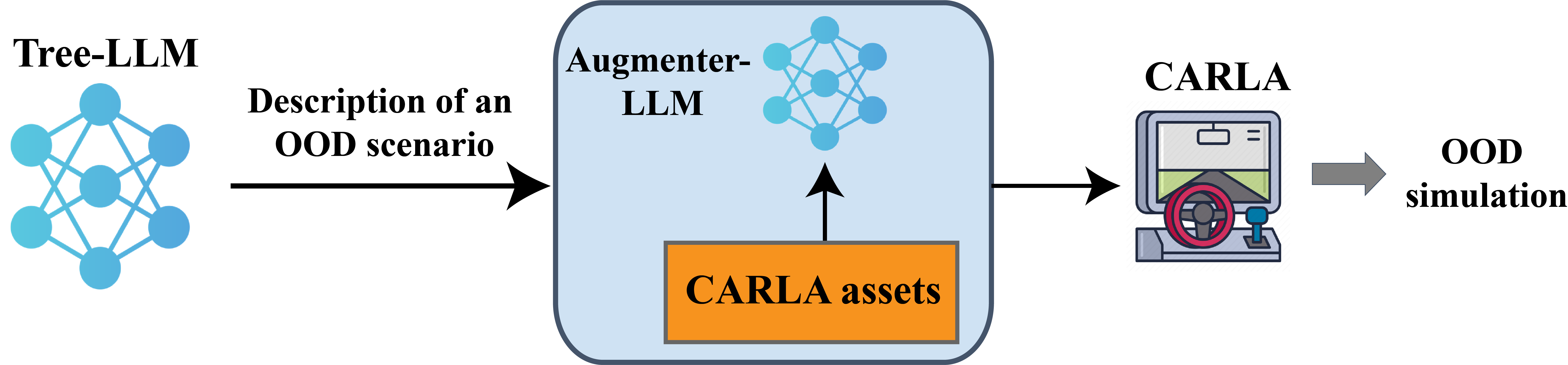}}
\caption{Automating the simulation of OOD scenarios in CARLA. The tree-LLM employs the simulatable tree to generate the textual description of an OOD scenario. The augmenter-LLM leverages CARLA assets to determine a scene configuration that aligns with the description of the OOD scenario.}
\label{fig:carla_automation}
\end{figure}

%%%%%%%%%%%%%%%%%%%%%%%%%%%%%%%%%%%%%%%%%%%%%%%%%%%%%%%%%%%%%%%%%%%%%%
%%%%%%%%%%%%%%%%%%%%%%%% METRICS %%%%%%%%%%%%%%%%%%%%%%%%%%%%%%%%%%%%%
\section{Metrics} \label{sec:metrics}

To evaluate the effectiveness of our framework in generating diverse OOD scenarios, we utilize two metrics:

\subsubsection{OOD-ness} \label{subsubsec:oodness}
To quantify the extent to which our generated scenarios are OOD, we employ a baseline of common urban driving behaviors and compare our generated data against it.  Let the baseline dataset, consisting of textual descriptions of typical urban traffic scenarios (e.g., nuScenes), be represented as $\mathcal{T}_b = \{t_{b,i}\}_{i=1}^{N}$, where $t_{b,i}$ represents the textual description of the $i^{\text{th}}$ sample, and $N$ is the total number of baseline descriptions. Similarly, let our generated dataset, containing textual descriptions of OOD scenarios, be denoted as $\mathcal{T}_o = \{t_{o,j}\}_{j=1}^{M}$, with $M$ being the total number of OOD samples.
We introduce an \textit{OOD-ness} metric to measure how far our generated samples deviate from the baseline, by evaluating their similarity within the contextual embeddings space of LLMs. Specifically, we use the textual embeddings of the samples in our dataset and measure their similarity with those of the baseline samples. Using a textual encoder $F$, we translate the textual descriptions in the baseline into their corresponding embeddings: $\forall i: e_{b,i} = F(t_{b,i})$, resulting in a dataset of baseline scenario embeddings: $\mathcal{D}_b = \{e_{b,i}\}_{i=1}^{N}$. Applying the same technique to our generated OOD dataset, we obtain a set of OOD scenario embeddings: $\mathcal{D}_o = \{e_{o,j}\}_{j=1}^{M}$. For each sample $j$, we calculate its OOD-ness by:
\begin{equation}
    ood_{o,j}^{b} = \min_i \frac{e_{b,i}^\top \, e_{o,j}}{||e_{b,i}|| \, ||e_{o,j}||},
    \label{eqn:oodness}
\end{equation}
which reflects the dissimilarity of sample $j$ relative to the baseline samples, based on the cosine similarity of their textual embeddings (see Fig.~\ref{fig:oodness}). To determine the overall OOD-ness of our dataset compared to the baseline, we compute the mean OOD-ness of all samples in our dataset:
\begin{equation}
    ood_{o}^{b} = mean([ood_{o,1}^{b}, ood_{o,2}^{b}, ..., ood_{o,M}^{b}]) 
\end{equation}

This metric effectively captures the distance between the textual embeddings of our generated samples and those in the baseline. A higher $ood_{o}^{b}$ value indicates that the samples in dataset $\mathcal{T}_o$ are more out-of-distribution compared to the baseline dataset $\mathcal{T}_b$.

\subsubsection{Diversity}
\begin{figure} [htb]
    \centering
    \subfigure[OOD-ness]
    {\includegraphics[width=0.45\columnwidth]{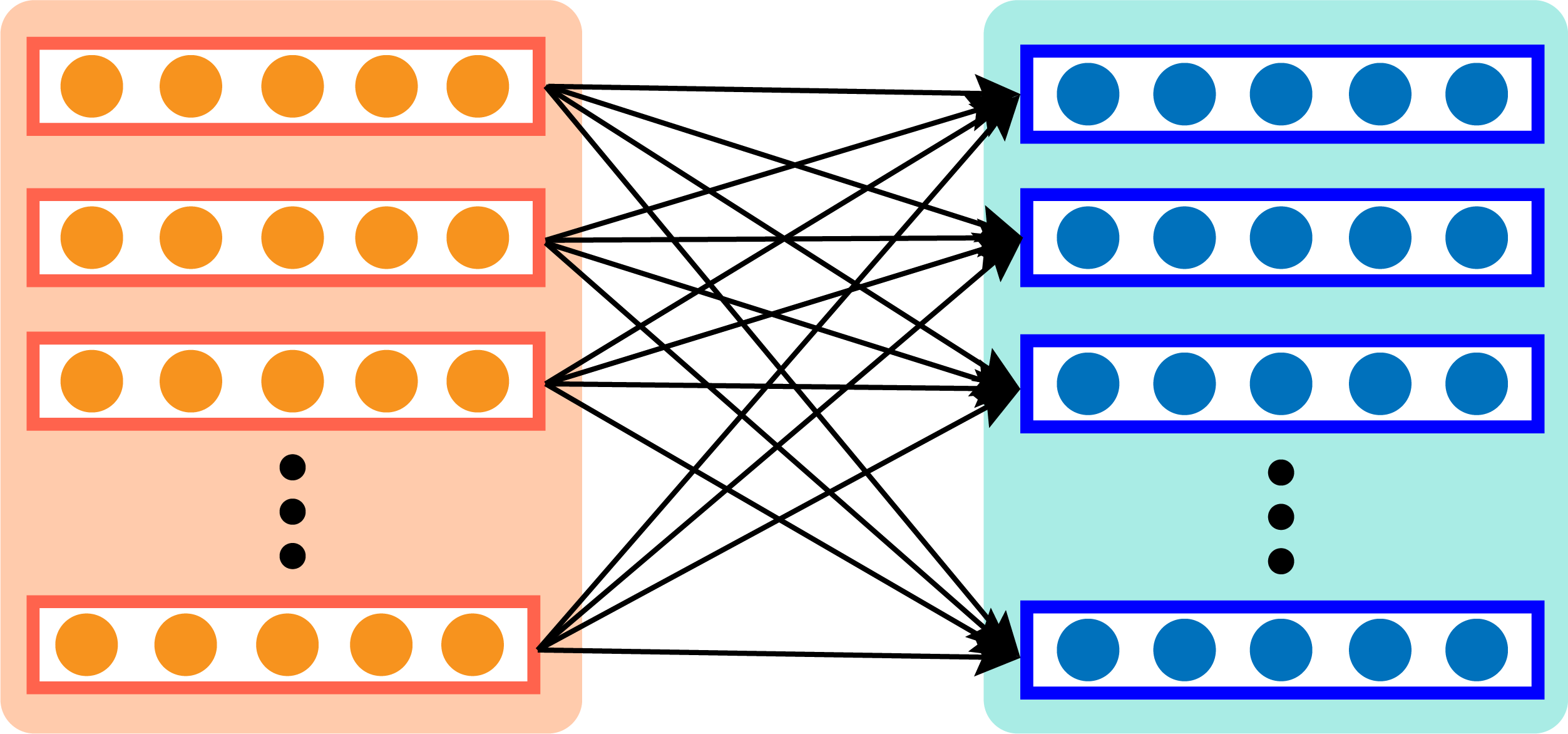}\label{fig:oodness}}
    \subfigure[Diversity]
    {\includegraphics[width=0.50\columnwidth]{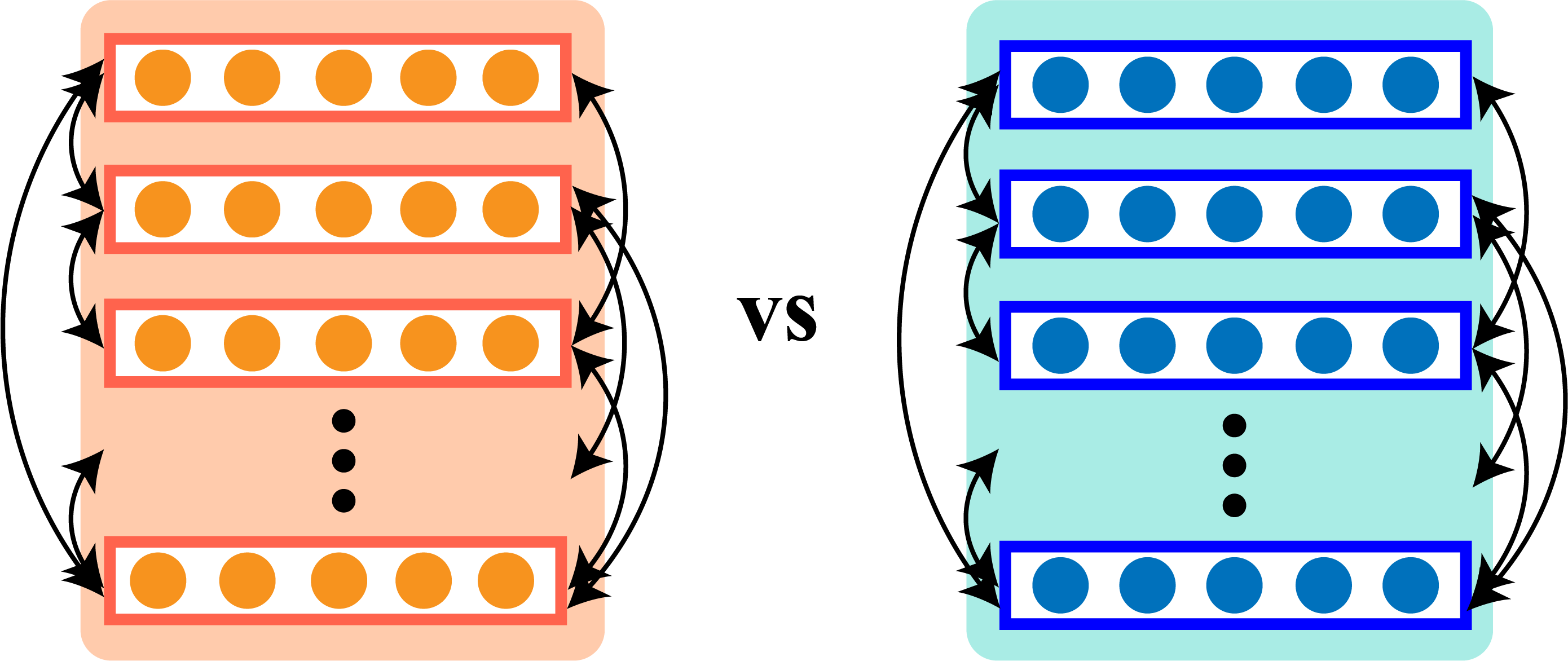}\label{fig:diversity}}
    \caption{Evaluation metrics used to assess the performance of our framework. The orange box represents our dataset, while the blue box represents a baseline consisting of common urban driving scenarios. (a) OOD-ness, calculated by comparing the similarity of textual embeddings from our samples to those in the baseline, (b) Diversity, determined by measuring the self-similarity scores of the textual embeddings within each dataset and comparing them against each other.}
    \vspace{-2mm}
\label{fig:evaluation_metrics}
\end{figure}

To evaluate the diversity of our generated OOD scenarios, we calculate the self-similarity of our dataset and compare it with the baseline of common urban driving scenarios (see Fig.~\ref{fig:diversity}). For each sample $j$ in our dataset, we determine the self-similarity score by finding the maximum cosine similarity between the embedding of sample $j$ and the embeddings of all other samples in the dataset:
\begin{equation}
    sim_{o,j} = \max_{\substack{j' \in \{1,...,M\}\\j' \neq j}} \frac{e_{o,j'}^\top \, e_{o,j}}{||e_{o,j'}|| \, ||e_{o,j}||}
\end{equation}

This score reflects the similarity of sample $j$ to the other samples in our dataset. The overall self-similarity score of our dataset is then computed as the average of these individual scores:
\begin{equation}
    sim_o = mean([sim_{o,1}, sim_{o,2}, ..., sim_{o,M}])
\end{equation}

Similarly, we can compute the self-similarity score for the baseline, denoted by $sim_b$. To quantify the diversity of a dataset, we define the diversity score as $div = - sim$, which indicates the extent to which the samples differ from one another. By comparing the diversity scores of our dataset and the baseline, we gain valuable insights into the diversity of the samples in each dataset.

%%%%%%%%%%%%%%%%%%%%%%%%%%%%%%%%%%%%%%%%%%%%%%%%%%%%%%%%%%%%%%%%%%%%%%
%%%%%%%%%%%%%%%%%%%%%%%% RESULTS %%%%%%%%%%%%%%%%%%%%%%%%%%%%%%%%%%%%%
\section{RESULTS}
\label{section:results}

We assess the effectiveness of our framework by: 1) creating textual datasets of OOD scenarios in different sizes, and 2) simulating a range of OOD scenarios in CARLA. These scenarios are subsequently employed to evaluate the performance of VLMs in managing OOD situations. The GPT-4o model serves as the LLM in our framework, and all of the simulations are conducted in CARLA 0.9.9.4. As the baseline for computing OOD-ness and diversity metrics, we utilize the nuScenes textual dataset \cite{caesar2020nuscenes}, as applied in the DriveLM paper \cite{sima2023drivelm}. For computing the textual embeddings of the scenarios, we use the 'all-MiniLM-L6-v2' sentence transformer.

\subsection{Diverse OOD Scenario Generation} \label{subsec:results_diverse_generation}
Following the structure of Fig.~\ref{fig:initial_tree}, we prompt GPT-4o with a few-shot chain-of-thought approach, to achieve an initial tree with 40 nodes. The structure of the initial tree is later refined by applying the red teaming technique (see Fig.~\ref{fig:diverse_tree}) for 100 iterations, to achieve a diverse tree with 77 nodes. Examples of the OOD scenarios generated by the diverse are as following:
\textit{
\begin{itemize}
    \item "The ego vehicle is moving through an area where a construction project is taking place. A piece of heavy equipment has been accidentally left out of place, creating an unmarked construction zone with barricades positioned incorrectly.",
    \item "The ego vehicle is moving through a city when a sudden traffic management system failure during rush hour causes all traffic lights and electronic signals in a busy urban area to become non-functional.",
    \item "The ego vehicle is moving on a road when a sudden police checkpoint appears on the road."
\end{itemize}
}

We use the diverse tree to generate textual datasets of OOD scenarios with 10, 100, and 1000 samples. The OOD-ness scores for these datasets are reported in the second row of Table~\ref{table:oodness-diversity}. To assess the values reported in this row, we randomly selected 100 samples from the baseline dataset and calculated their OOD-ness relative to the rest of the baseline. By repeating this process for 10 times and averaging the results, we obtained an OOD-ness score of -0.953 for a batch of 100 baseline samples. Comparing this score with the values in the second row of Table~\ref{table:oodness-diversity} indicates that our generated scenarios are notably OOD compared to the typical urban driving baselines.

Similarly, we calculate the diversity scores for our OOD datasets with sample sizes of 10, 100, and 1000. The results, along with the diversity score of the nuScenes baseline, are presented in the third row of Table~\ref{table:oodness-diversity}. The higher diversity scores of our datasets compared to the baseline highlight the robustness of our framework in generating a wide range of OOD scenarios.

\begin{table}[htb]
\footnotesize
    \centering
    \begin{tabular}{|p{1.5cm}|p{0.9cm}|p{1.2cm}|p{1.2cm}|p{1.5cm}|}
        \hline
        \centering Dataset  & \centering baseline & \centering ours-10 & \centering ours-100 & \makecell{\centering ours-1000} \\
        \hline
        \centering OOD-ness & \centering NA & \centering -0.677 & \centering -0.691 & \makecell{\centering -0.690} \\
        \hline
        \centering Diversity & \centering -0.781 & \centering -0.635 & \centering -0.638 & \makecell{\centering -0.642} \\
        \hline
    \end{tabular}
    \caption{}
    \label{table:oodness-diversity}
    \vspace{-4mm}
\end{table}

\begin{remark}
    Given that the diverse tree is constructed using LLMs, and considering the vast knowledge these models possess, some of the OOD scenarios generated by the diverse tree might represent exceedingly rare events. 
    % For instance, "\textit{Ego is moving through a busy city street when a personal aircraft makes an emergency landing on the road}". 
    Although such scenarios are still theoretically possible, their likelihood is very minimal. To address this concern and maintain a balanced level of OOD-ness in the generated scenarios, we can set an upper-bound threshold on the acceptable OOD-ness score of the dataset and exclude any scenarios that exceed this limit. Similarly, we can adjust a lower-bound threshold to ensure that the generated scenarios are not too similar to typical urban driving scenarios.
\end{remark}

\subsection{Automated Simulation of OOD Scenarios} \label{subsec:results_automated_simulation}

\begin{figure*}[htb]
\centerline{\includegraphics[width=0.90\textwidth]{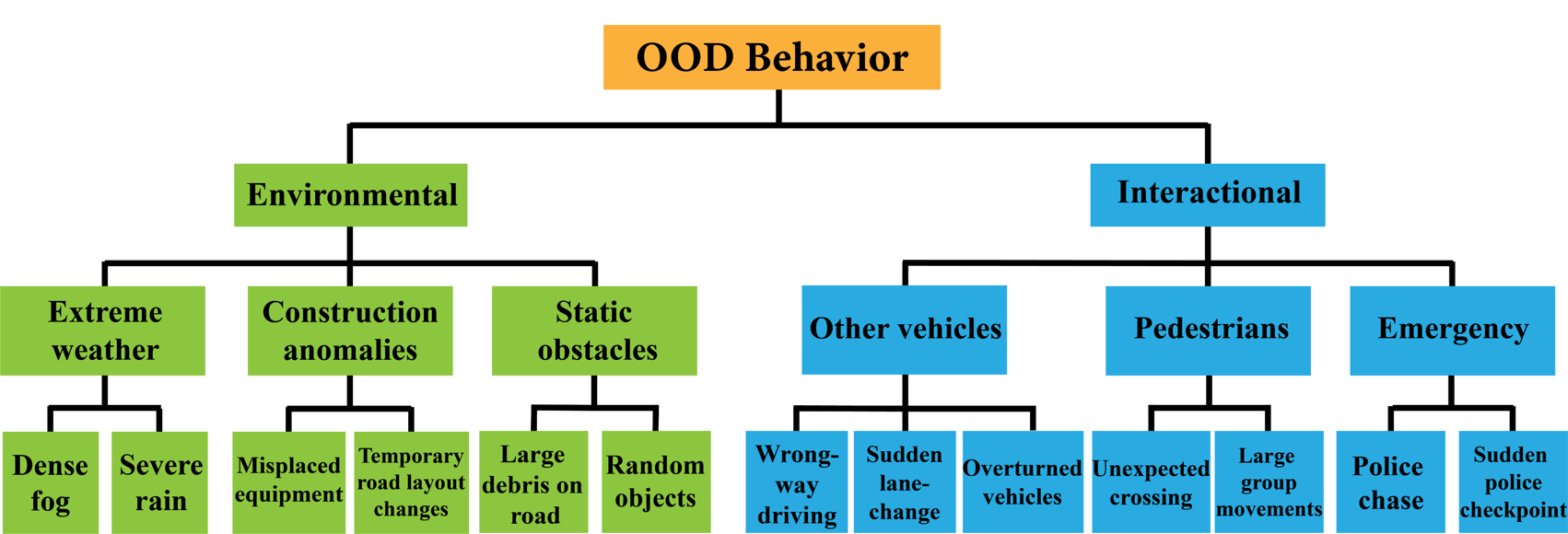}}
\caption{Illustration of the simulatable tree, developed by pruning the diverse tree based on the available assets in CARLA.}
\label{fig:tree_example}
\end{figure*}

In the initial phase of automating the simulation of OOD scenarios, we refine the diverse tree into a simulatable tree, tailored to the available assets in CARLA. Fig.~\ref{fig:tree_example} depicts the simulatable tree we have developed based on CARLA, which contains 22 nodes, in contrast to the 77 nodes of the original diverse tree. The tree-LLM utilizes the simulatable tree to generate textual descriptions of simulatable OOD scenarios, by tracing paths from the root to the leaves. The simulatable tree comprises 13 unique paths, and we prompt the tree-LLM to generate 10 scenarios along each path, resulting in a textual dataset of 130 OOD scenarios. Examples of the generated OOD scenarios are as following:
\begin{itemize}
    \item "\textit{Ego is traveling on a two-lane road at night in thick fog, with another vehicle ahead in the same lane. The visibility is greatly diminished due to the fog, making it hard to see far ahead.}" (see Fig.~\ref{fig:dense_fog}),
    \item "\textit{The ego vehicle is traveling on a two-lane residential road under clear and sunny weather conditions. A large cardboard box, serving as static debris, is positioned in the middle of the road ahead.}",
    \item "\textit{Ego is moving on a three-lane road under clear weather conditions, with an overturned vehicle positioned ahead in the same lane.}"
\end{itemize}

The OOD-ness score of our simulatable dataset is -0.765. Compared to the OOD-ness score of a random batch of 100 samples from the baseline, which is -0.953, it is clear that our simulatable scenarios are promisingly OOD. Moreover, the diversity score of generated dataset is -0.762. Compared to the diversity of the baseline in Table~\ref{table:oodness-diversity}, our simulatable dataset still has higher scenario richness than the baseline. The decrease in the diversity score of scenarios generated by the simulatable tree, compared to those generated by the diverse tree, is attributed to the pruning process used to construct the simulatable tree from the diverse tree, which restricts the range of scenario variability.

\begin{figure} [htb]
    \centering
    \subfigure[]
    {\includegraphics[width=0.49\columnwidth]{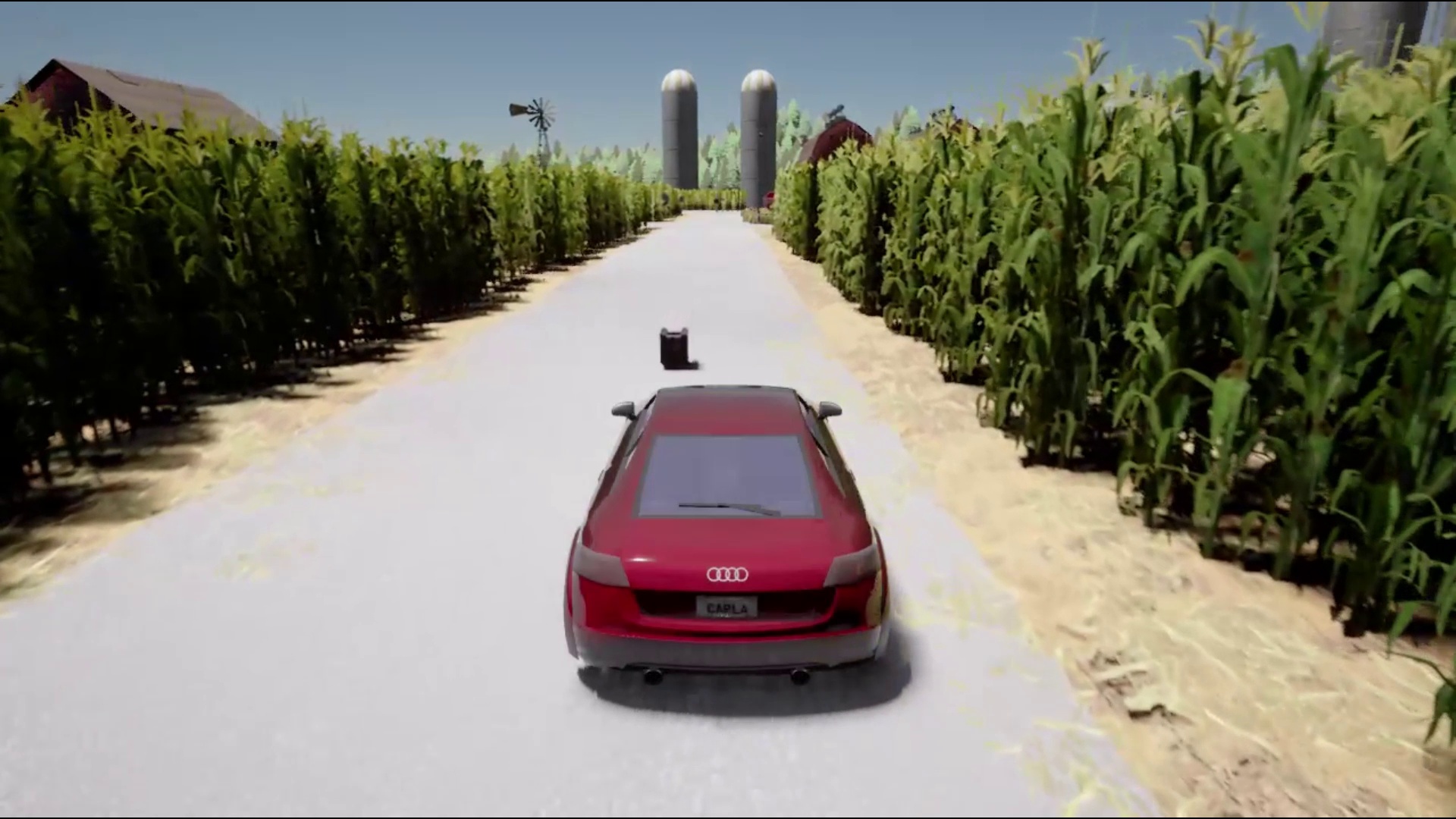}\label{fig:random}}
    \subfigure[]
    {\includegraphics[width=0.49\columnwidth]{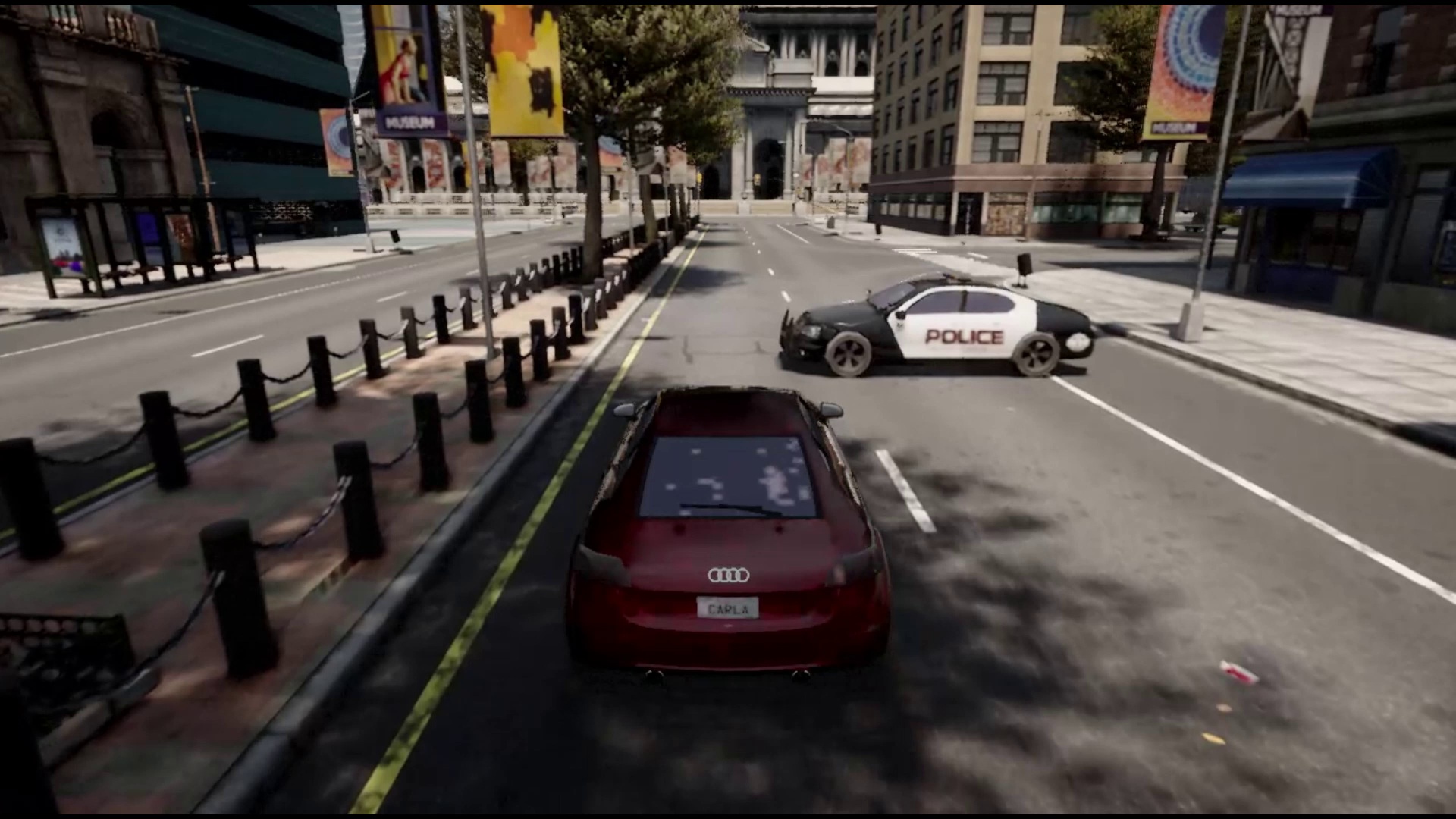}\label{fig:police}}
    \subfigure[]
    {\includegraphics[width=0.49\columnwidth]{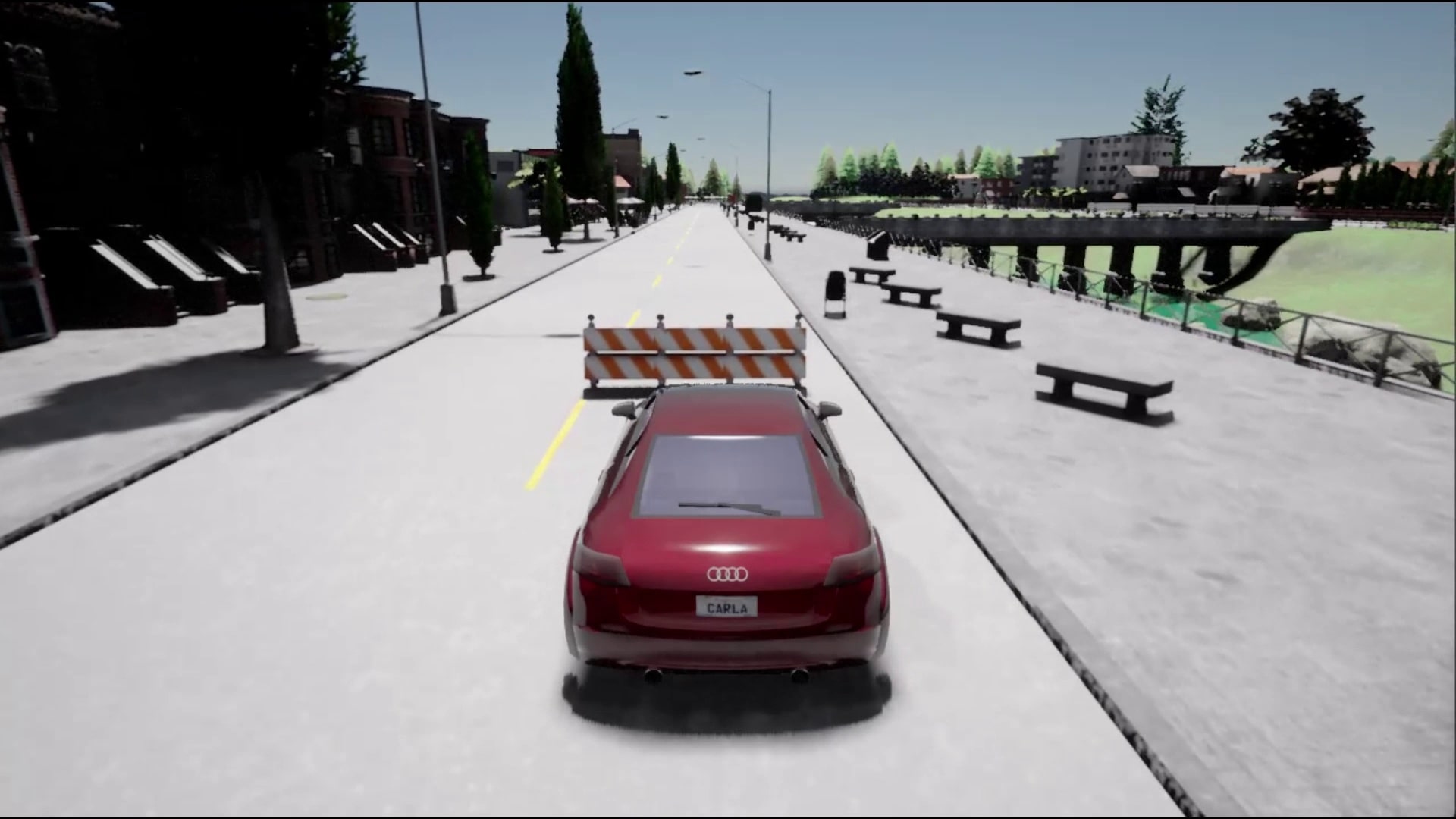}\label{fig:temporary}}
    \subfigure[]
    {\includegraphics[width=0.49\columnwidth]{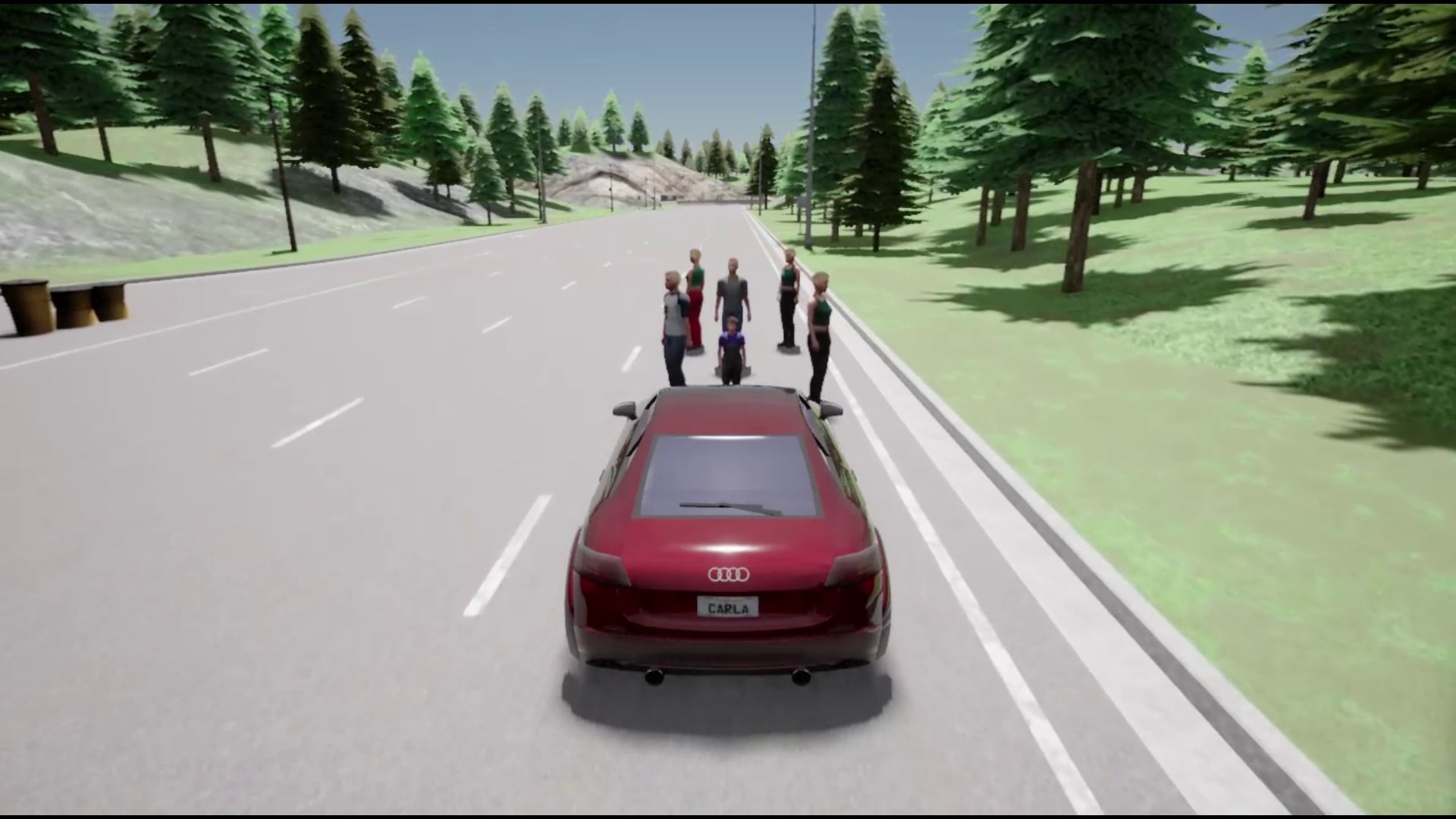}\label{fig:group}}
    \caption{Examples of the OOD scenarios that are automatically simulated in CARLA: (a) a random object (plant pot) on the road, (b) a sudden police checkpoint along the ego's lane, (c) temporary change in the layout of the road, (d) a large group of people on the road (parade, protest, etc).}
\label{fig:simulation_examples}
\end{figure}

Following the remainder of the pipeline for automating the simulation of OOD scenarios (see Fig.~\ref{fig:carla_automation}), the textual descriptions of the simulatable scenarios are fed into the augmenter-LLM. This model determines the specific details for simulating the scenarios, including weather conditions, attributes and relative state of other objects, and their behavior models. The generated simulation details are then automatically parsed and applied into the simulator to generate the corresponding scenario. Examples of the simulated OOD scenarios are shown in Fig.~\ref{fig:ood_examples} and~\ref{fig:simulation_examples}.

As evidenced by the evaluation metrics and generated examples, our proposed framework shows strong potential for generating OOD scenarios in autonomous driving. It's important to note that expanding the range of simulatable scenarios can be achieved by developing bigger diverse tree and enriching the assets and libraries within the simulator. This, in turn, results in a richer simulatable tree and, consequently, OOD scenarios.

\subsection{Evaluating the Performance of VLMs} \label{subsec:results_vlm_performance}

In the final stage of our study, we assess the performance of state-of-the-art VLMs on our simulated data, focusing on two primary tasks: identifying the OOD nature of the generated scenarios, and selecting the appropriate safe control action to navigate through them. This study is highly significant as it demonstrates the capability of language models to address OOD scenarios in autonomous driving. The VLMs used in this part include GPT-4o, Claude 3.5 Sonnet \cite{claude3}, and Gemini 1.5 Pro \cite{gemini}. As a preprocessing step, we extract three essential frames from each simulation trace—the first, middle, and last—for scenario analysis. As we process each scenario, we present these frames sequentially to the VLMs, prompting them in a zero-shot CoT approach to describe each frame. These descriptions serve as background information for both of the following tasks.

\begin{figure}[htb]
\centerline{\includegraphics[width=0.97\columnwidth]{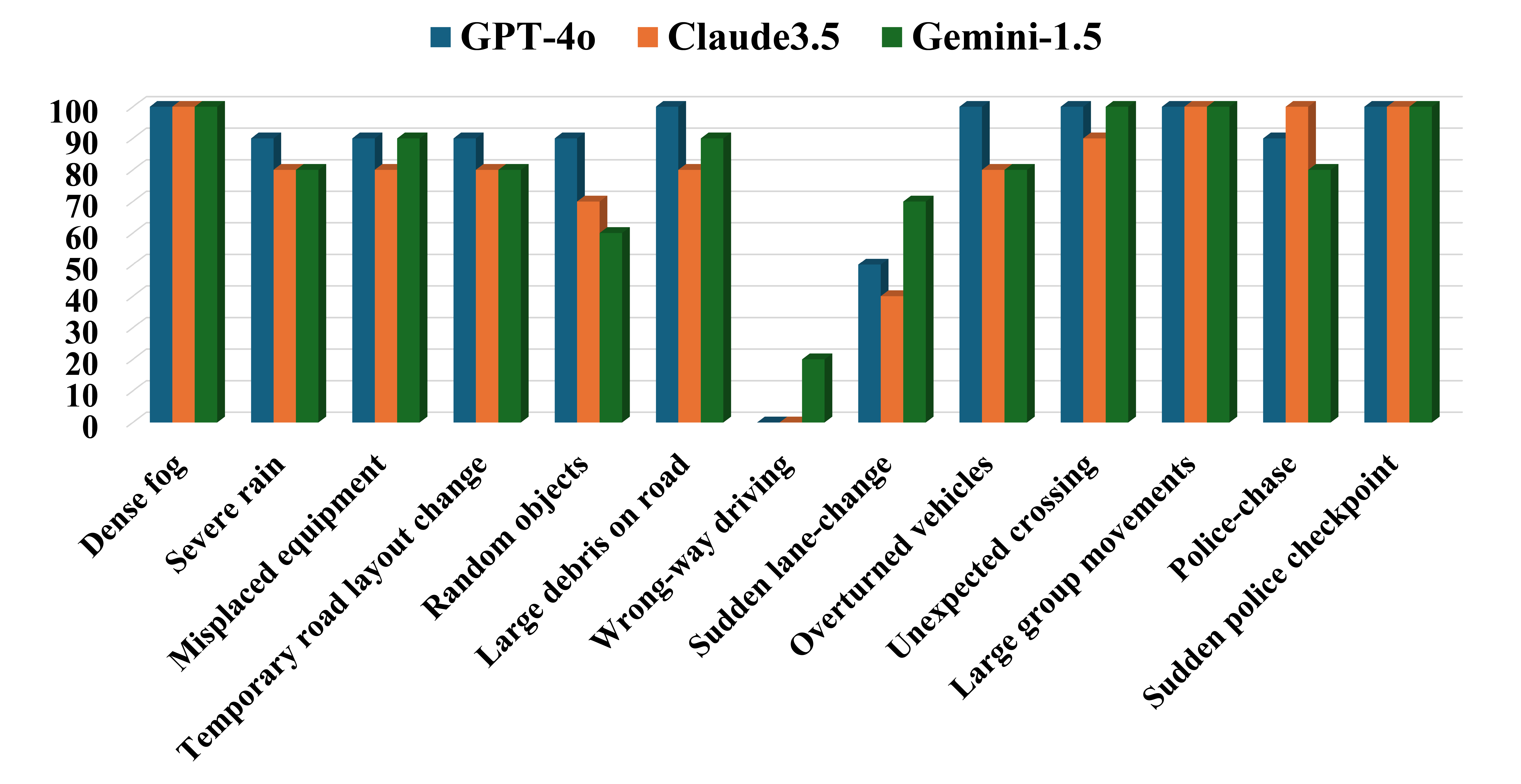}}
\caption{Success rate of VLMs in inferring the OOD-ness of the scenarios.}
\label{fig:vlm_performance_ood}
\end{figure}

\begin{figure}[htb]
\centerline{\includegraphics[width=0.95\columnwidth]{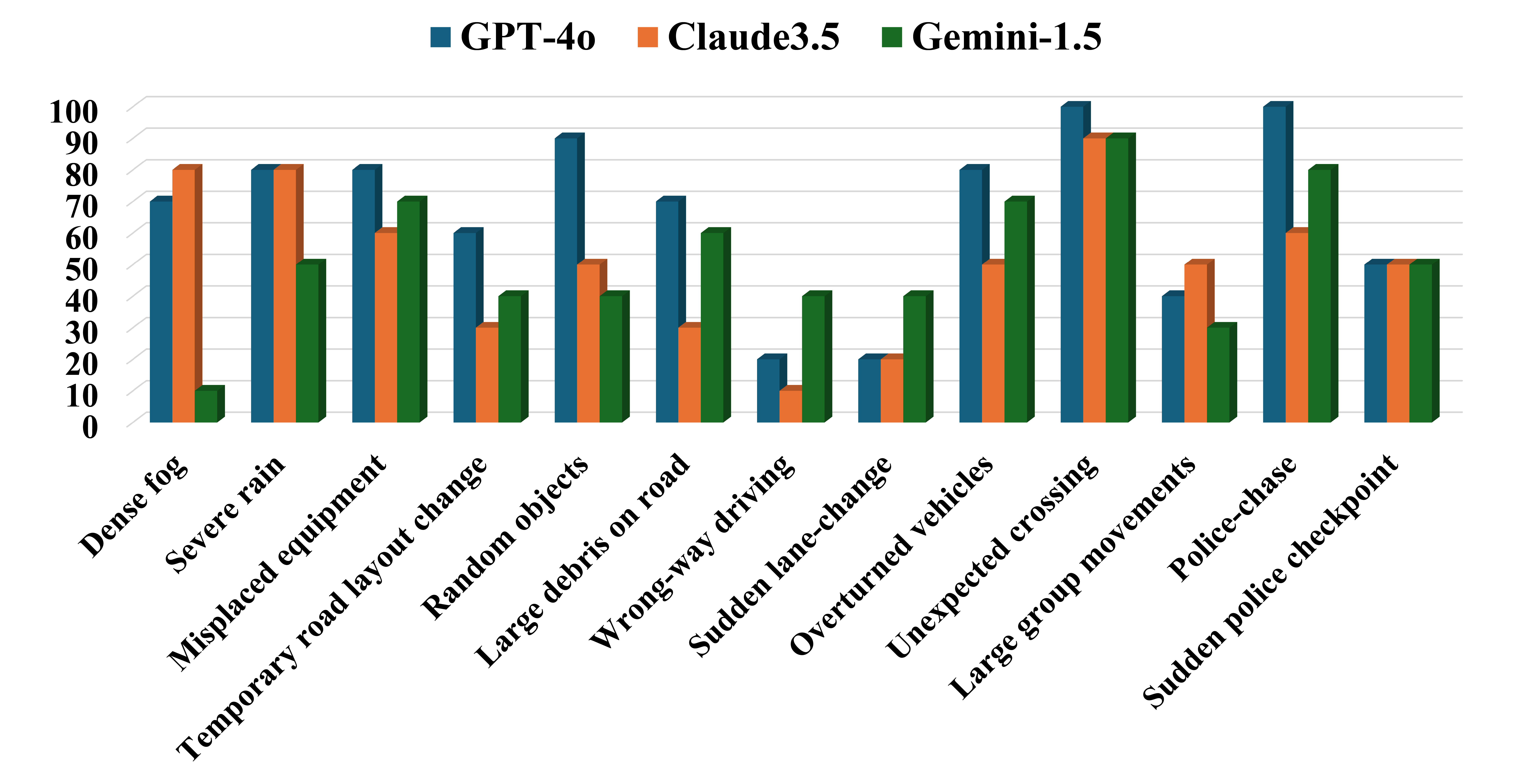}}
\caption{Success rate of VLMs in choosing a safe control for each scenario.}
\label{fig:vlm_performance_control}
\end{figure}

In the first task, we query the VLMs to determine the OOD characteristics of the generated scenarios, comparing their outputs with the ground-truth labels. The results, presented in Fig.~\ref{fig:vlm_performance_ood}, show the success rate of the models in accurately identifying the OOD nature of each scenario. As it is clear from the figure, the VLMs generally perform well in accurately inferring the OOD-ness of the scenarios, with the exception of scenarios involving another vehicle driving the wrong way or making a sudden lane change, where their performance—particularly in the wrong-way driving case—is significantly lacking. 
In the second task, we query the VLMs to identify a safe control action (in terms of collision avoidance) to take in each scenario. The VLMs are presented with a set of discrete actions (e.g., change to the right lane, change to the left lane, keep moving forward, etc.) to choose from, and each action is mapped to a control sequence to be executed by ego in CARLA. Fig.~\ref{fig:vlm_performance_control} shows the success rate of the models in selecting a safe control action in each scenario. The results indicate that none of the VLMs demonstrate strong zero-shot performance in this task; however, GPT-4o performs relatively better than the other models. Our findings align with with existing work \cite{sreeram2024probing}, which explored the use of language models as world models for driving.

Table~\ref{table:vlm_performance} summarizes the average success rates over all the scenarios: the second row provides the average success rate of the VLMs in identifying the correct OOD aspect of the scenarios, while the third row shows the average success rate in selecting a safe control action. The table underscores that while the VLMs show reasonable performance in identifying the OOD nature of the scenarios, their ability to determine a safe control action remains unreliable, with GPT-4o emerging as the best-performing model among them.

\begin{table}[htb]
\scriptsize
    \centering
    \begin{tabular}{|p{2.0cm}|p{0.9cm}|p{1.9cm}|p{1.9cm}|}
        \hline
        \centering Model  & \centering GPT-4o & \centering Claude 3.5 Sonnet & \makecell{\centering Gemini 1.5 Pro}\\
        \hline
        \centering OOD-ness & \centering 84.61 \% & \centering 76.92 \% & \makecell{\centering 80.76 \%}\\
        \hline
        \centering Safe control & \centering 66.15 \% & \centering 50.77 \% & \makecell{\centering 51.53 \%}\\
        \hline
    \end{tabular}
    \caption{}
    \label{table:vlm_performance}
    \vspace{-4mm}
\end{table}

%%%%%%%%%%%%%%%%%%%%%%%%%%%%%%%%%%%%%%%%%%%%%%%%%%%%%%%%%%%%%%%%%%%%%%
%%%%%%%%%%%%%%%%%%%%%%%%%%%%%% CONCLUSION %%%%%%%%%%%%%%%%%%%%%%%%%%%%
\section{CONCLUSION} \label{section:conclusion}
In this paper, we present a framework for generating OOD scenarios in autonomous driving. Our method constructs a diverse tree to generate OOD cases and automatically simulates them using CARLA. Our approach shows promising performance in generating OOD scenarios, evaluated by OOD-ness and diversity metrics, and compared with a common urban driving baseline. Furthermore, we evaluate the performance of the modern VLMs in identifying the OOD nature of our generated scenarios and selecting safe control actions to address them. Among the tested VLMs, GPT-4o demonstrates relatively better performance compared to the others.

%%%%%%%%%%%%%%%%%%%%%%%%%%%%%%%%%%%%%%%%%%%%%%%%%%%%%%%%%%%%%%%%%%%%%%
%%%%%%%%%%%%%%%%%%%%%%%%%%%%%% REFERENCES %%%%%%%%%%%%%%%%%%%%%%%%%%%%
\bibliographystyle{IEEEtran}
\bibliography{references}

\end{document}